\documentclass{article}
\usepackage{spconf,amsmath,epsfig,amsmath}
\usepackage{booktabs} 
\usepackage{times}
\usepackage{epsfig}
\usepackage{graphicx}
\usepackage{subfigure}
\usepackage{amsmath}
\usepackage{amssymb}
\usepackage{hyperref}
\usepackage{color}
\usepackage{multirow}
\usepackage{pgfplots}
\usepackage{adjustbox}
\usepackage{amsfonts}
\usepackage{tikz}
\usepackage{enumitem}
\usepackage{wrapfig}

\begin{document}\sloppy

\def\x{{\mathbf x}}
\def\L{{\cal L}}

\title{GeoCapsNet: Aerial to Ground view Image Geo-localization using Capsule Network}
%
\name{Bin Sun$^1$, Chen Chen$^2$, Yingying Zhu$^1$, Jianmin Jiang$^1$}
\address{$1.$ Shenzhen University, China, 518060\\
$2.$ Department of Electrical and Computer Engineering, University of North Carolina at Charlotte\\
2161230410@email.szu.edu.cn, chen.chen@uncc.edu, \{zhuyy, jianmin.jiang\}@szu.edu.cn}

\maketitle

\begin{abstract}
The task of cross-view image geo-localization aims to determine the geo-location (GPS coordinates) of a query ground-view image by matching it with the GPS-tagged aerial (satellite) images in a reference dataset. Due to the dramatic changes of viewpoint, matching the cross-view images is challenging. In this paper, we propose the GeoCapsNet based on the capsule network for ground-to-aerial image geo-localization. The network first extracts features from both ground-view and aerial images via standard convolution layers and the capsule layers further encode the features to model the spatial feature hierarchies and enhance the representation power. Moreover, we introduce a simple and effective weighted soft-margin triplet loss with online batch hard sample mining, which can greatly improve the image retrieval accuracy. Experimental results show that our GeoCapsNet significantly outperforms the state-of-the-art approaches on two benchmark datasets. \textit{The source code will be released soon}.
\end{abstract}

\section{Introduction}
\label{sec:intro}
Image geo-localization refers to the problem of determining where (i.e. GPS coordinates) an image is taken from based on the visual information only. This research has attracted widespread attention in recent years, due to its potential applications in autonomous driving, augmented reality, to name a few. Traditional geo-localization approaches requires accurate telemetry and sensor model e.g. digital ortho-quad (DOQ) ~\cite{int3}, and Digital Elevation Map (DEM) to perform geo-registration with the reference data. However, these accurate models are difficult to obtain. Recently, geo-localization based on image matching has attracted growing interest since it is free from the constraint of requiring the meta data ~\cite{hays2008im2gps,zamir2010accurate}. 
A typical solution to this problem relies on matching ``ground-to-ground" using a reference database of geo-tagged photographs ~\cite{hays2008im2gps,torii2011visual,zamir2014image,shan2014accurate,schindler2007city}. These methods are relatively easy because both query and reference images are ground-level and they are in the same domain. One main drawback of such approaches is that the reference dataset contains geo-tagged images which are concentrated in cities and tourist attractions. However, ground-level images of some geo-graphical locations may not have geo-location information. Therefore, ground-level image matching based methods cannot scale to global scale due to lack of reference data. 

\begin{figure}[t]
\center
	\includegraphics[width=0.35\textwidth]{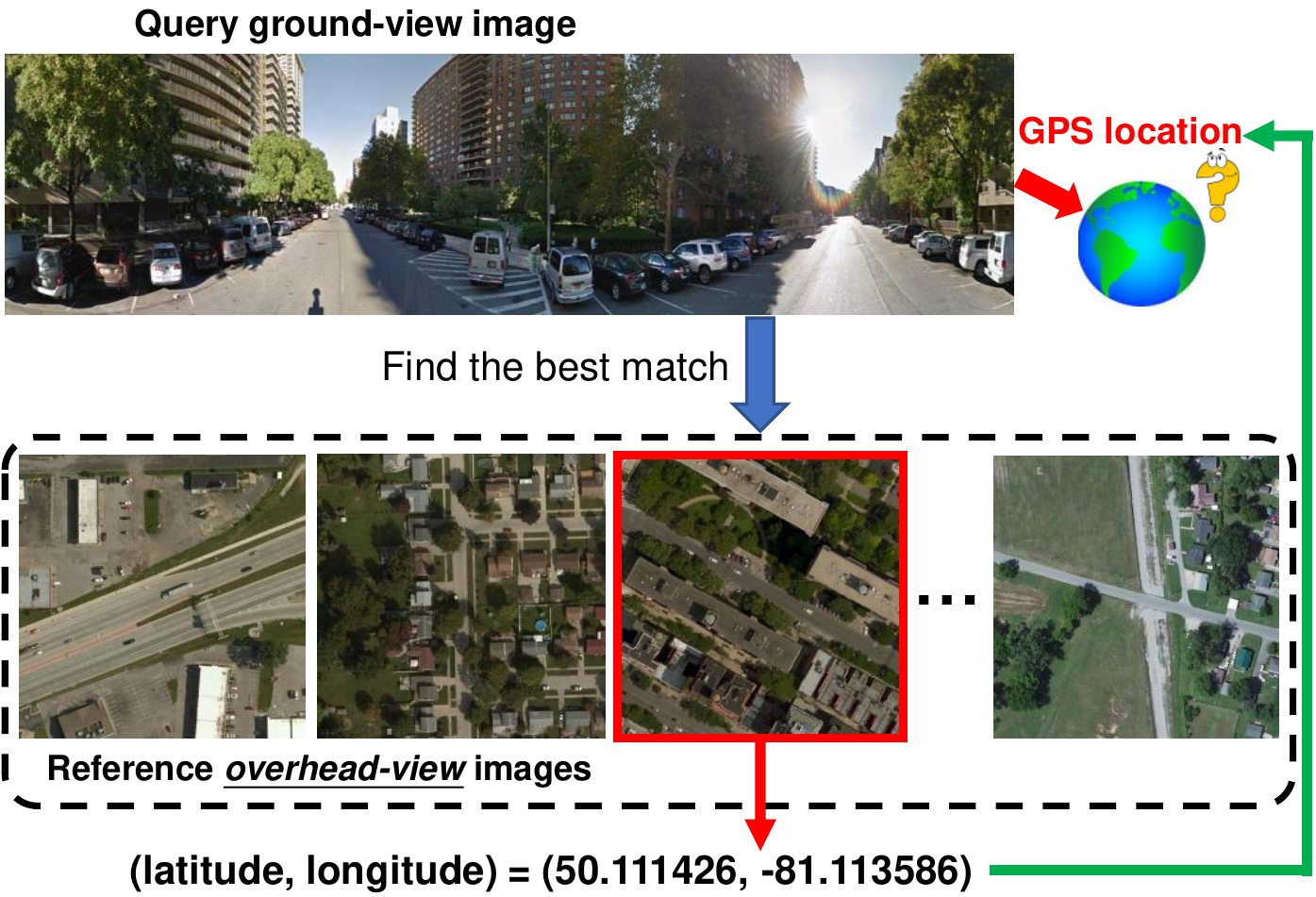}
	\caption{Cross-view image geo-localization. Given a street-view image as a query, the goal of geo-localization is to determine its GPS location by matching it with a reference database of overhead satellite images with GPS coordinates. Due to viewpoint difference, the visual contents look very different in cross-view images. }
	\label{fig:figure1}
\end{figure}

On the other hand, thanks to the advent of satellite and aerospace surveys, aerial photographs densely cover the entire planet. As a result, matching ground-level photos to aerial imagery (e.g. Google satellite imagery) has become an attractive alternative to the geo-localization problem ~\cite{bansal2016ultrawide,lin2013cross,shan2014accurate,lin2015learning,workman2015location,workman2015wide,vo2016localizing,stumm2016robust,Zhai2017Predicting,tian2017cross,Hu_2018_CVPR}. 
As shown in Fig. \ref{fig:figure1}, cross-view image matching is a very challenging task because of the drastic change in viewpoint between ground and aerial images. A key element of cross-view image matching is to learn the powerful feature presentation of the cross-view images, such that the distance of a matched pair of images is small whereas the distance of the unmatched pair is large in this feature space. However, learning feature embedding using
pairs of images only considers the visual information similarity. The geometric discrepancy between two cross-view images is not properly addressed. 

\textbf{Motivation.} 
Recently, capsule network has been proposed ~\cite{sabour2017dynamic} to address some of the limitations of convolutional neural networks (CNN). 
Capsule network enables building parts-to-whole relationship between entities and allows capsules to learn viewpoint invariant representations. Inspired by these properties of the capsule network, we propose an aerial to ground view image geo-localization approach, namely GeoCapsNet, by leveraging the feature representation power of the capsule network. It tasks as input cross-view (ground view and overhead view) image pairs, matched or unmatched, and learns a feature embedding space such that features of the matching image pairs are close
and unmatched image pairs are far apart. The contributions of the paper are:
\vspace{-0.1in}
\begin{itemize}[leftmargin=*]
\item 
We propose an end-to-end network architecture GeoCapsNet for cross-view image-based geo-localization. Our work expands the use of capsule
network to the task of image matching for the first time. 
\vspace{-0.1in}
\item
We introduce a new weighted soft-margin triplet loss with online hard sample mining in each training batch. We show the batch hard-mining process is effective for improving the generalization ability of the network and therefore can boost the image retrieval performance.
\item
\vspace{-0.1in}
Extensive experiments on two datasets demonstrate that our GeoCapsNet significantly outperforms the state-of-the-art algorithms for cross-view image geo-localization.
\end{itemize}

\section{Related work}
In this section, we provide a review of the state-of-the-art solutions to the cross-view image geo-localization problem. Lin et al. \cite{lin2013cross} introduced the ``discriminative translation'' approach in which an aerial image classifier is trained based on ground-level scene matches for ground-to-overhead geo-localization. Bansal et al. \cite{bansal2016ultrawide} matched query street-level facades to airborne imagery under viewpoint and illumination variation by selecting the intrinsic facade motif scale and modeling facade structure through self-similarity. Shan et al. \cite{shan2014accurate} proposed a fully automated ground-based multi-view stereo model for matching ground-level photos to aerial imagery, which is capable of handling drastic viewpoint variations by adopting a novel view-dependent feature matching approach. 
Workman et al. \cite{workman2015wide} used multi-scale overhead images for the same location in order to perform cross-view training by embedding the feature representations from both views in a joint semantic feature space. Vo et al. \cite{vo2016localizing} introduced a distance-based logistic loss to improve the performance of Siamese network and Triplet network for cross-view geo-localization. They also showed that explicit orientation supervision can improve the localization accuracy. Zhai et al. \cite{Zhai2017Predicting} developed a new network architecture to predict semantic layout of ground-level images from the corresponding overhead images, which can also be used for several other tasks such as orientation estimation and geo-calibration. Hu et al. \cite{Hu_2018_CVPR} adopted the fully convolutional network to extract local image features, which are then encoded into global image descriptors using the NetVLAD \cite{netvlad}. 
Although a great deal of effort has been devoted to build discriminative feature representations for cross-view images, it still remains challenging for cross-view image matching due to the large differences in visual contents and scene structures.

\section{PROPOSED GEOCAPSNET}

\subsection{Capsule Network}

\begin{figure*}[t]
	\centering
	\includegraphics[width=0.7\textwidth]{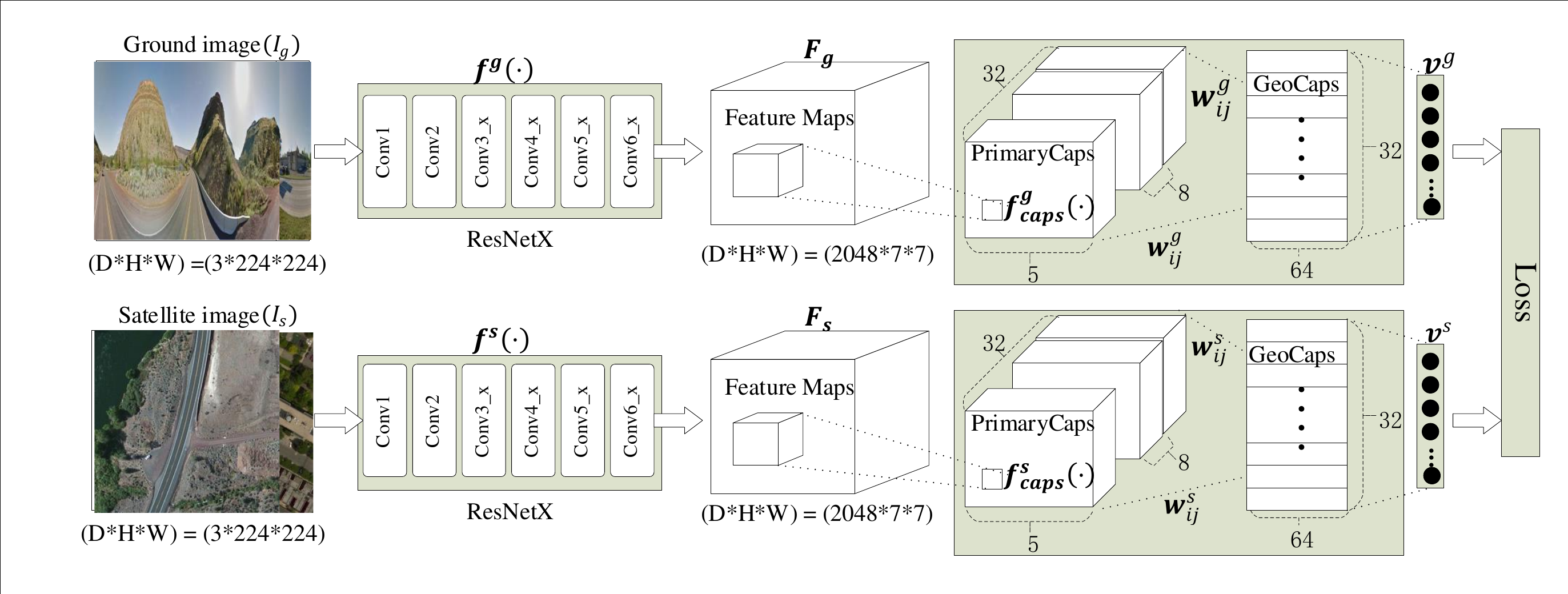}
	\vspace{-0.1in}
	\caption{The architecture of GeoCapsNet, which is a two-branch Siamese network takes as input a pair of cross-view images. Each network branch consists of two parts: ResNetX and Capsule layers (PrimaryCaps and GeoCaps layers).}
	\label{fig:geocaps}
\end{figure*}

The capsule network~\cite{sabour2017dynamic} uses a group of neurons to represent an entity. The important information about the state of the features detected by all capsules in the capsule network is encapsulated in the form of a vector. Since the neurons in the traditional network layers are too simple to represent a concept, the capsule network uses vectors as feature representations in the capsule layers. The output vector of the capsule 
represents two parts: (1) its length represents the probability of occurrence of an instance (e.g. object, visual concept or part thereof), (2) its direction indicates graphical properties of the object (e.g. position, color, direction, shape, etc.)

In the cross-view image matching problem, aerial and ground images share some semantics, e.g. road, tree, building, etc. Moreover, the scene layout and geometric structure 
are also important cues for image matching. Inspired by the capsule networks' capability of modeling spatial relationships (i.e. orientation and position) of extracted features, we propose an end-to-end cross-view image matching network incorporating the capsule layers, dubbed as GeoCapsNet, to encode the relative spatial relationship between features to obtain a powerful image representation. In the following section, we present the details of GeoCapsNet. 

\subsection{GeoCapsNet Architecture}
The overall architecture of the proposed GeoCapsNet is shown in Fig. \ref{fig:geocaps}. It follows the Siamese network \cite{chopra2005learning} structure with two identical networks in parallel. The input to the two networks is the ground and satellite image, respectively. For higher-level capsules to obtain semantic representations, we begin with a residual network structure called ResNetX to extract the semantic features of images. Following the convention of ResNet \cite{he2016deep}, the details of the ResNetX structure are in Table \ref{tab:resnetx}. It consists of two convolutional layers and four residual blocks (i.e. Conv3\_x - Conv6\_x). ResNetX uses Batch Normalization at every layer. The max-pooling layer is not used to preserve the information about the input data.

\begin{table}[th]
\footnotesize
	\begin{center}
		\begin{tabular}{c|c|c}
			\hline
			Layer name & Output size & Layer 
			\\
			\hline
			Conv1 & $112\times112$ &  $7\times7$, 64, stride 2 \\
			\hline
			Conv2  & $56\times56$  &  $3\times3$, 64, stride 2 \\
			\hline
			Conv3\_x &  $56\times56$ &
			$ 
			\begin{bmatrix}1\times1,&64\\3\times3,&64\\1\times1,&256\end{bmatrix}\times3
			$ 
			\\
			\hline
			Conv4\_x &  $28\times28$ &
			$ 
			\begin{bmatrix}1\times1,&128\\3\times3,&128\\1\times1,&256\end{bmatrix}\times4
			$ 
			 \\
			\hline
			Conv5\_x &  $14\times14$ & 
			$ 
			\begin{bmatrix}1\times1,&256\\3\times3,&256\\1\times1,&1024\end{bmatrix}\times6
			$  
			\\
			\hline
			Conv6\_x & $7\times7$ & 
			$ 
			\begin{bmatrix}1\times1,&512\\3\times3,&512\\1\times1,&2048\end{bmatrix}\times3
			$  
			\\
			\hline
		\end{tabular}
		\vspace{-0.05in}
		\caption{The structure of ResNetX.}
		 \label{tab:resnetx}
	\end{center}
\end{table}

The output of ResNetX is 2048 feature maps with spatial size 7$\times$7, which are served as input to the capsule layer. We use two layers of capsules: PrimaryCaps and GeoCaps. 
The PrimaryCaps layer has 32 primary capsules whose job is to take basic features detected by the ResNetX and produce combinations of the features. The ``primary capsules" are very similar to convolutional layer in their nature~\cite{sabour2017dynamic}. Each capsule applies eight 3$\times$3$\times$2048 convolutional kernels (with stride 1) to the 7$\times$7$\times$2048 input volume and therefore produces 5$\times$5$\times$8 output tensor. Since there are 32 such capsules, the output volume has shape of 5$\times$5$\times$8$\times$32. The GeoCaps layer has 32 capsules, one for an entity in image. Each capsule takes as input a 5$\times$5$\times$8$\times$32 tensor i.e. 5$\times$5$\times$32 8-dimensional vectors.  As per the dynamic routing algorithm~\cite{sabour2017dynamic}, each of these input vectors gets their own 8$\times$64 weight matrix that maps 8-dimensional input space to the 64-dimensional capsule output space. The 32$\times$64-dimensional vector representation of an image is obtained.

Concretely, let $I_g$ and $I_s$ denote the ground and satellite image, respectively. $f^g(\cdot)$ and $f^s(\cdot)$ indicate the corresponding ResNetX structure for $I_g$ and $I_s$. In other words, the ResNetX models for $I_g$ and $I_s$ have separate model weights. The resulting features are $F_g=f^g (I_g)$ and $F_s=f^s (I_s)$ as shown in Fig. \ref{fig:geocaps}.
$F_g$ and $F_s$ are passed to the PrimaryCaps layer of each branch, generating the output vectors of each capsule:
$\boldsymbol u_i^s=f_{caps}^s(F_s),\;\boldsymbol u_i^g=f_{caps}^g(F_g)$. Then $\boldsymbol u_i^s$ and $\boldsymbol u_i^g$ are fed into the corresponding GeoCaps layer through the dynamic routing algorithm, and the output of each capsule is $\boldsymbol v_j^x$, where $x\in\left\{s,\;g\right\}$, and
\begin{equation}
\small
\widehat{\boldsymbol u}_{j\vert i}^x=\boldsymbol w_{ij}^x\boldsymbol u_i^x,\;\;\boldsymbol s_j^x=\sum_ic_{ij}^x\widehat{\boldsymbol u}_{j\vert i}^x
\nonumber,\;\;
\boldsymbol v_j^x=\frac{\left\|\boldsymbol s_j^x\right\|^2}{1+\left\|\boldsymbol s_j^x\right\|^2}\frac{\boldsymbol s_j^x}{\left\|\boldsymbol s_j^x\right\|}
\end{equation}
$\boldsymbol w_{ij}^x$ is a weight matrix that needs to be learned, $c_{ij}$ are coupling coefficients that are determined by the iterative dynamic routing process. The representation of an image can be formulated as $\boldsymbol v^x=\{\boldsymbol v_1^x,\boldsymbol v_2^x,\boldsymbol v_3^x,\cdots\boldsymbol v_k^x\}$, where $k$ represents the number of capsules in GeoCaps layer and $x\in\left\{s,\;g\right\}$ indicates the satellite or ground branch.

According to the weights learning strategy for the capsule layers in the two branches, we develop two variants of GeoCapsNet. Specifically, we denote two capsule branches with different model weights, i.e. $f_{caps}^s(\cdot) \neq f_{caps}^g(\cdot)$, as \textbf{GeoCapsNet-I}, and two capsule branches sharing the same model weights, i.e. $f_{caps}^s(\cdot)=f_{caps}^g(\cdot),\;\boldsymbol w_{ij}^s=\boldsymbol w_{ij}^g,\;c_{ij}^s=c_{ij}^g$, as \textbf{GeoCapsNet-II}.

\section{OBJECTIVE FUNCTION}
In image retrieval tasks, Contrastive loss ~\cite{varior2016gated}, Triplet loss ~\cite{schroff2015facenet,cheng2016person}, and Quadruplet loss ~\cite{chen2017beyond} are popular loss functions to train the deep neural networks. 
In image geo-localization, the goal of the loss function is to make the distance between the images of the same geo-location (positive pairs) as small as possible, and the distance between the images of different geo-locations (negative pairs) as large as possible. Take triplet loss as an example, a triplet is ensembled by randomly sampling three images from the training data, including an Anchor ($a$), a Positive sample ($p$) and a Negative sample ($n$). $(a,p)$ forms a positive pair and $(a,n)$ forms a negative pair. However, if the sample pairs are easy to distinguish, the network cannot learn a good feature representation, leading to poor generalization ability.
To this end, we introduce a batch-wise hard sample mining method.

\textbf{Batch construction and hard sample mining.} We select $M$ ground images in each training batch. For each ground image $a$ in the batch, its matching satellite image $p$ is used to construct the positive pair. Then a set of $M-1$ satellite images in different geo-locations as $a$ can be used to form negative pairs in the batch. This negative set of images is denoted as $B$. The triplet loss function is expressed as:
\begin{eqnarray}
L_{tri}=\frac1{M}\sum_{a\in batch}(d_{a,p}-\underset{n\in B}{min\;}d_{a,n}+\theta)_+
\end{eqnarray}

\begin{figure}[t]
	\centering
	\includegraphics[width=0.35\textwidth]{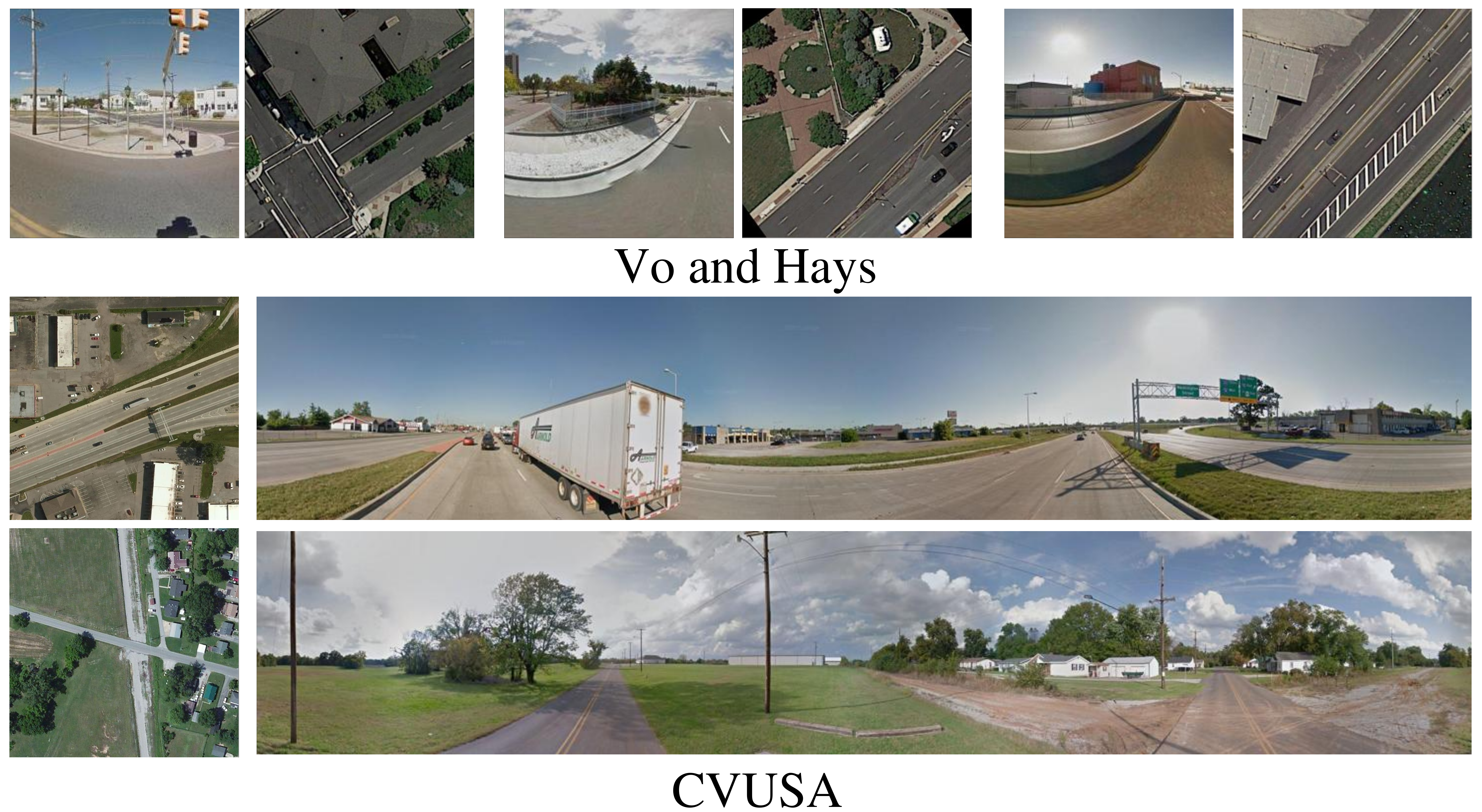}
	\vspace{-0.1in}
	\caption{Example cross-view images from two datasets.}
	\label{fig:dataset}
\end{figure}

As shown in Eq. 1, 
$d_{a,p}$ is the distance between the capsule feature of $a$ (i.e. $\mathbf{v}^g(a)$, see Fig. 2) and the capsule feature of $p$ (i.e. $\mathbf{v}^s(p)$).
$\underset{n\in B}{min\;}d_{a,n}$ finds the negative sample which is closest to $a$, i.e. the hardest sample in the batch, to calculate the triplet loss. $\theta \geq 0$ is the margin, and $(\cdot)_+ = max(0,\cdot)$.

To avoid manually setting the margin $\theta$, we adopt the soft-margin triplet loss~\cite{Hu_2018_CVPR}: $L_{soft}\;=\ln\left(1+e^d\right)$, where $d=d_{a,p}-d_{a,n}$. To improve the convergence rate, the weighted soft-margin ranking loss~\cite{Hu_2018_CVPR} scales $d$ in $L_{soft}$ by a coefficient $\alpha$: $L_{weighted}\;=ln\left(1+e^{\alpha d}\right)$. Therefore, our weighted soft-margin triplet loss with batch hard-mining ({\bf{Soft-TriHard Loss}}) can be expressed as:
\begin{eqnarray}
L_{sth}=\frac1{M}\sum_{a\in batch}\ln\left(1+e^{\alpha\left(d_{a,p}-\underset{n\in B}{min}d_{a,n}\right)}\right)
\end{eqnarray}

\section{Experimental Results}
\textbf{Datasets.} We evaluate our GeoCapsNet on two cross-view datasets - CVUSA~\cite{Zhai2017Predicting} and Vo and Hays~\cite{vo2016localizing}. The CVUSA consists of matching pairs of ground panoramas and satellite image. It contains 35532 image pairs for training and 8884 image pairs for testing. Vo and Hays dataset consists of street-view and overhead images from 11 different cities in the U.S. with more than 1 million pairs of images. Follow the same experimental setting in \cite{Hu_2018_CVPR}, we randomly select 9 cities, 8 of which are used to train our network, and the 9th -- Denver city is for testing. Fig. 3 shows a few examples from the datasets.

\noindent \textbf{Evaluation metric.} The models are evaluated by the recall accuracy at top 1\% for our networks, as is done in~\cite{Hu_2018_CVPR}.  The recall at top 1\% is percentage of cases in which the correct satellite match of the query ground view image is ranked within top 1 percentile.

\subsection{Implementation Details}
The proposed model is trained for 50 epochs on the training set with a batch size of 32 ($M=32$). We implement GeoCapsNet in Tensorflow. Adam optimizer~\cite{kinga2015method} is used for training the model with an initial learning rate of $10^{-3}$. RELU is the activation unit. Regularization is implemented in combination of L2-regularization and Batch Normalization. $\alpha$ is set to 15 in the Soft-TriHard loss.
In GeoCaps layer, we set the number of capsule to 32, and the number of dynamic routing iterations to 4. L2 normalization is applied to the last layer feature of GeoCapsNet.

\begin{table}[h]
\footnotesize
	\begin{center}
		\begin{tabular}{|c|c|c|}
			\hline
			
			&\multicolumn{2}{|c|}{Recall @top1\%}   \\
			\cline{2-3}

			& Vo and Hays~\cite{vo2016localizing} & CVUSA~\cite{Zhai2017Predicting}\\
			
			\hline
			Workman et.al~\cite{workman2015wide} & 15.40\%  & 34.30\% \\
			\hline
			Vo and Hays~\cite{vo2016localizing} &59.90\% & 63.70\% \\
			\hline
			Zhai et al~\cite{Zhai2017Predicting} &---- & 43.20\% \\
			\hline
			CVM-Net ~\cite{Hu_2018_CVPR}&67.90\% & 91.40\% \\
			\hline
			GeoCapsNet-I &69.59\% & 96.52\% \\
			\hline
			GeoCapsNet-II &\textbf{76.83\%} & \textbf{98.07\%} \\
			\hline
		\end{tabular}
		\caption{Performance comparison of our GeoCapsNet with the state-of-the-art cross-view geo-localization approaches. } \label{tab:recall}
	\end{center}
\end{table}

\subsection{Results and Ablation Study}
\textbf{Comparison to existing approaches.} We compare our proposed GeoCapsNets to four state-of-the-art methods \cite{workman2015wide,vo2016localizing,Zhai2017Predicting,Hu_2018_CVPR} on two datasets. 
Table \ref{tab:recall} shows the top 1\% accuracies of our GeoCapsNet and other methods. 
Both GeoCapsNet-I and GeoCapsNet-II outperform all the other approaches by considerable margins on two datasets, leading to new state-of-the-art results.
The results also reveal that GeoCapsNet-II achieves better performance than GeoCapsNet-I, suggesting that the weight sharing scheme of the two branch capsule layers forces the network to learn close and similar internal relationship representation between the cross-view images. 
It is also noted that all approaches have higher accuracy on CVUSA than Vo and Hays~\cite{vo2016localizing} dataset because the ground image in CVUSA is panoramic, which contains more information than the single-view image in Vo and Hays~\cite{vo2016localizing}.

\begin{figure}[t]
	\centering
	\includegraphics[width=0.35\textwidth]{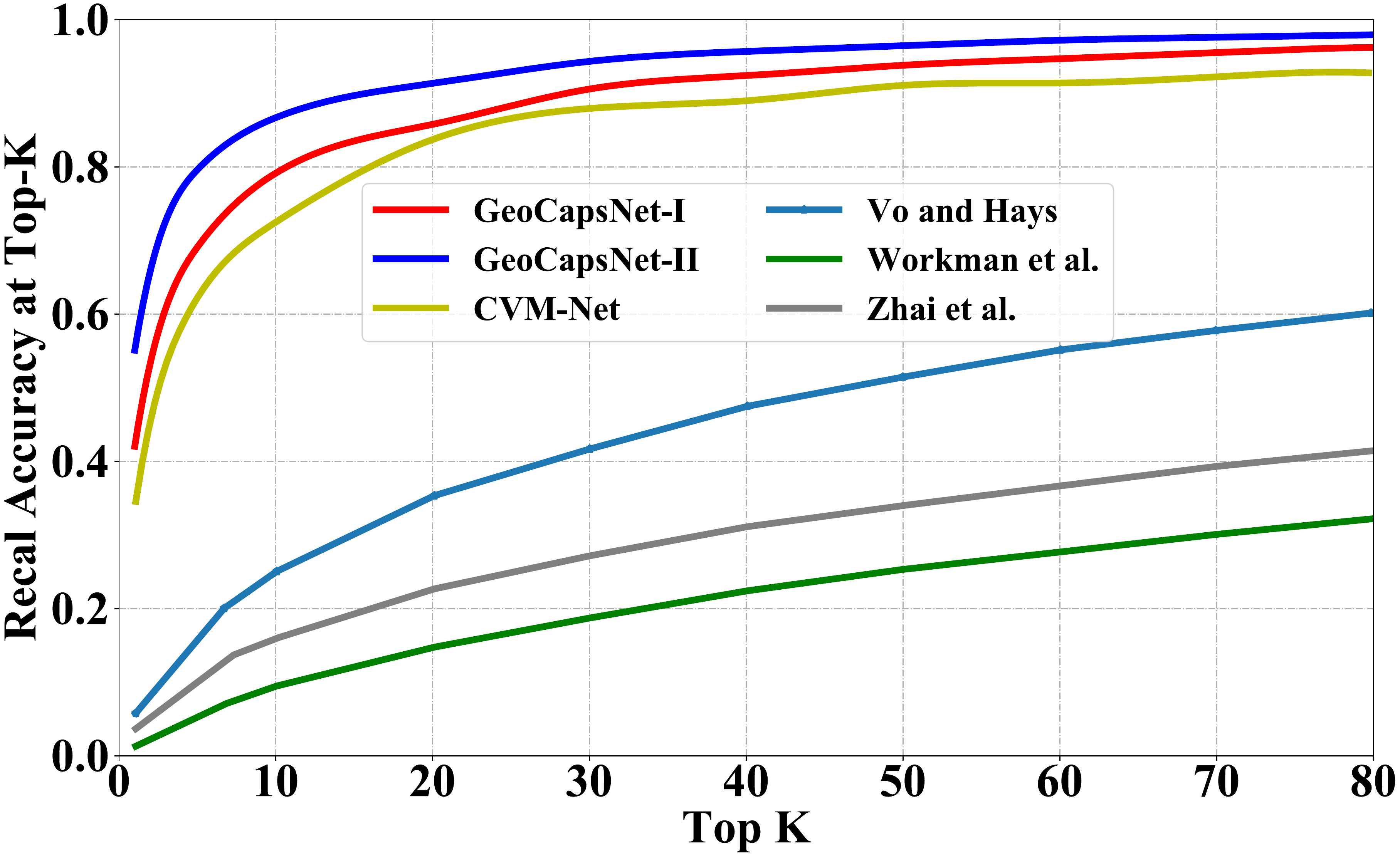}
	\vspace{-0.1in}
	\caption{Top-K recall accuracy on CVUSA.}
	\label{fig:top}
\end{figure}

Fig. \ref{fig:top} plots the Top-K (K=1--80) recall accuracy of our GeoCapsNets and other approaches on the CVUAS dataset. It is evident that our GeoCapsNets achieve much better performance than other approaches. The performance gain of GeoCapsNets over the state-of-the-art CVM-Net \cite{Hu_2018_CVPR} is significant especially in the range of Top-1 to Top-20 recall, e.g. 20.53\% improvement at Top-1 recall (GeoCapsNet-II 55.09\% \textit{vs} CVM-Net 34.56\%).

\textit{To understand the effectiveness of our proposed GeoCapsNet, we conduct several ablation experiments to investigate the contribution of each important component.}

\begin{wrapfigure}{r}{0.25\textwidth}
    \centering
    \vspace{-0.1in}
    \includegraphics[width=0.25\textwidth]{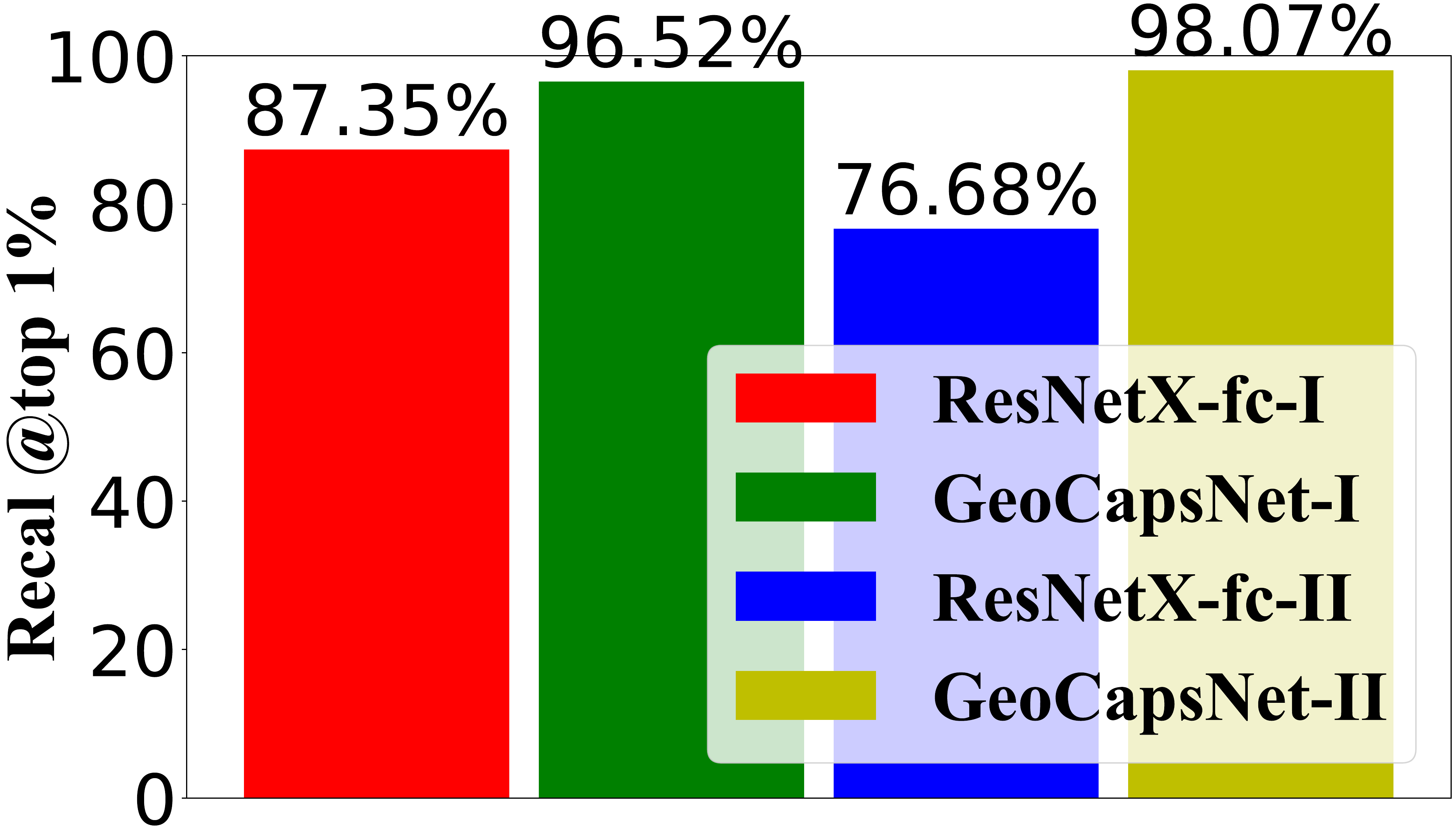}
    \vspace{-0.2in}
    \caption{\small Recall accuracy of GeoCapsNet without capsule layers on CVUSA.}
    \label{fig:ab_capsule}
\end{wrapfigure}

\textbf{Capsule layers.} In this experiment, we replace the capsule layers (PrimaryCaps and GeoCaps layer) in our GeoCapsNets with a fully connected layer as the final feature representation to form the ResNetX-fc-I and ResNetX-fc-II networks. The proposed soft-TriHard loss is used. The results in Fig. \ref{fig:ab_capsule} clearly demonstrate the advantage of using capsule layers for encoding more discriminative features.

\begin{table}[h]
	\small
	\begin{center}
		\begin{tabular}{|c|c|c|c|c|}
			\hline
			&Triplet&Soft-TriHard\\
			\hline
			GeoCapsNet-I & 70.38\% & 96.52\%\\
			\hline
			GeoCapsNet-II &77.46\% & 98.07\%\\
			\hline
			
		\end{tabular}
		\caption{Recall@top1\% of GeoCapsNet with different losses.} \label{tab:hard}
	\end{center}
\end{table}

\textbf{Batch hard-mining.} To demonstrate the effectiveness of the batch hard sample mining procedure, we remove this process in our \textit{Soft-TriHard} loss function. Therefore, the loss reduces to the weighted soft-margin triplet loss ~\cite{Hu_2018_CVPR}, represented by \textit{Triplet} in Table \ref{tab:hard}. The results on the CVUSA datast suggest that batch hard-mining is very effective and is able to significantly boost the performance.

\begin{figure}[t]
	\centering
	\includegraphics[width=0.35\textwidth]{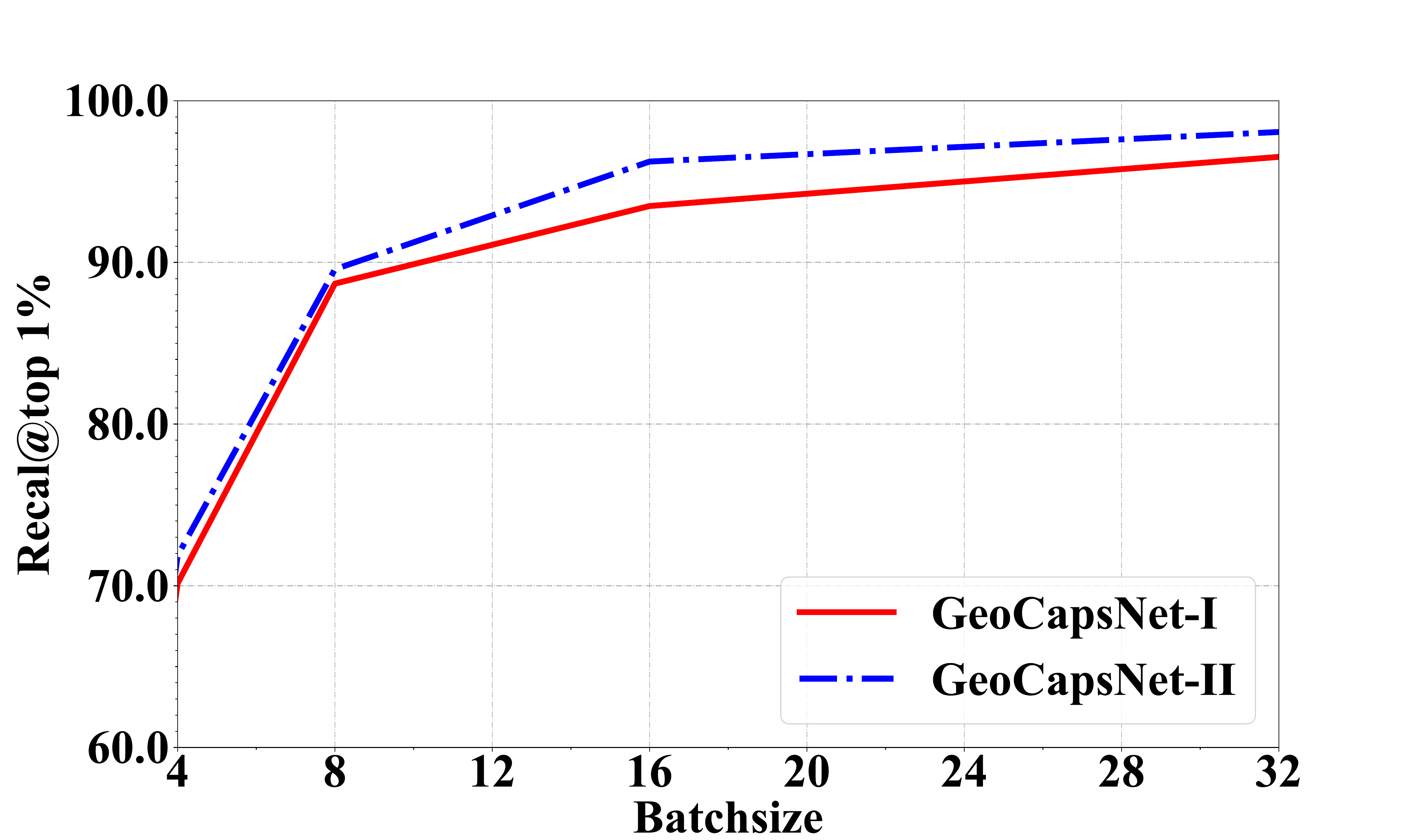}
	\caption{Performance of our GeoCapsNets on CVUSA
		with different batch sizes.}
	\label{fig:batchsize}
\end{figure}

\textbf{Batch size.} As described in Section 4, we select $M$ ground images in each training batch and construct the positive and negative pairs. We analyze the performance of our GeoCapsNets with different batch sizes. Specifically, we tune different values of batch size while keep other parameters the same. In Figure \ref{fig:batchsize}, as the batch size increases, the Top 1\% recall accuracy of our GeoCapsNets becomes higher. This is because in our batch-hard mining method, the larger the batch size, the larger the search range of the samples, so that harder samples can be obtained. To balance the performance and memory requirement, we set $M=32$.

\begin{figure*}[t]
	\centering
	\includegraphics[width=0.78\textwidth]{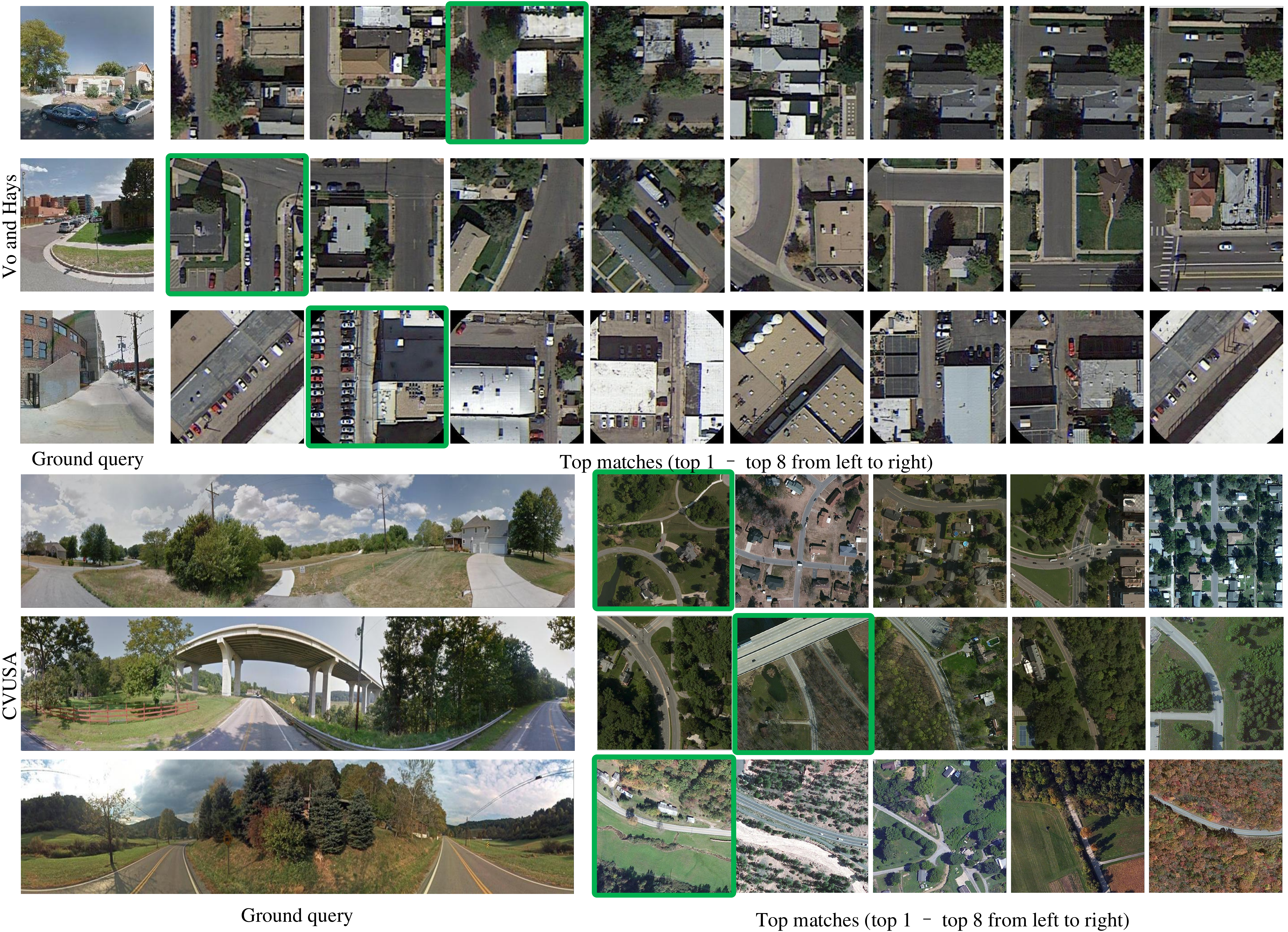}
	\vspace{-0.1in}
	\caption{Image retrieval examples of GeoCapsNet on two datasets. The image marked by the green box is the ground truth.}
	\label{fig:result}
\end{figure*}

\textbf{Model comparison.} 
Table \ref{tab:model} provides a model comparison between  GeoCapsNet and CVM-Net in terms of the parameter size of the network, the storage size of the model, and the length of the feature encoding for image retrieval. The number of parameters in our GeoCapsNet is much smaller than that of the CVM-Net, leading to a more compact model. 
In addition, the code length (i.e. feature dimension) of GeoCapsNet is only half the length of CVM-Net. In image retrieval, the shorter length of image feature coding means less computational complexity and faster retrieval speed. 
Finally, we show a few examples of geo-localizing query ground-view images using our GeoCapsNet in Fig. \ref{fig:result}. \textbf{Please refer to the supplementary material for more examples and analysis.}

\begin{table}[h]
	\footnotesize
	\begin{center}
		\begin{tabular}{|c|c|c|c|}
			\hline
				 &  \# Parameters & Model size &Code length\\
			\hline
			CVM-Net \cite{Hu_2018_CVPR} & 160,311,424  & 1.8G &4096\\
			\hline
			GeoCapsNet-I & 82,764,672 & 947.51M &2048\\
			\hline
			GeoCapsNet-II & \textbf{64,938,624} & \textbf{743.51M} &2048\\
			\hline
		\end{tabular}
		\caption{Comparison of GeoCapsNets and CVM-Net ~\cite{Hu_2018_CVPR}.} \label{tab:model}
	\end{center}
\end{table}

\section{Conclusion}
In this paper, we presented a cross-view image geo-localization method by matching query ground images with geo-tagged reference satellite images. We proposed the GeoCapsNet architecture which captures high-level semantic features of images and their relationships due to the capsule layers. An effective batch hard sample mining is incorporated into the weighted soft-margin ranking loss, which greatly improves the retrieval accuracy of our network. Our approach significantly outperforms the state-of-the-art methods on two large-scale datasets.

\bibliographystyle{IEEEbib}
\small{\bibliography{icme2019template}}

\begin{thebibliography}{10}

\bibitem{int3}
Barbara Zitova and Jan Flusser,
\newblock ``Image registration methods: a survey,''
\newblock {\em Image and vision computing}, 2003.

\bibitem{hays2008im2gps}
James Hays and Alexei~A Efros,
\newblock ``Im2gps: estimating geographic information from a single image,''
\newblock in {\em CVPR}, 2008.

\bibitem{zamir2010accurate}
Amir~Roshan Zamir and Mubarak Shah,
\newblock ``Accurate image localization based on google maps street view,''
\newblock in {\em ECCV}, 2010.

\bibitem{torii2011visual}
Akihiko Torii, Josef Sivic, and Tomas Pajdla,
\newblock ``Visual localization by linear combination of image descriptors,''
\newblock in {\em ICCV Workshops}, 2011.

\bibitem{zamir2014image}
Amir~Roshan Zamir and Mubarak Shah,
\newblock ``Image geo-localization based on multiplenearest neighbor feature
  matching usinggeneralized graphs,''
\newblock {\em IEEE TPAMI}, 2014.

\bibitem{shan2014accurate}
Qi~Shan, Changchang Wu, Brian Curless, Yasutaka Furukawa, Carlos Hernandez, and
  Steven~M Seitz,
\newblock ``Accurate geo-registration by ground-to-aerial image matching,''
\newblock in {\em 3D Vision}, 2014.

\bibitem{schindler2007city}
Grant Schindler, Matthew Brown, and Richard Szeliski,
\newblock ``City-scale location recognition,''
\newblock in {\em CVPR}, 2007.

\bibitem{bansal2016ultrawide}
Mayank Bansal, Kostas Daniilidis, and Harpreet Sawhney,
\newblock ``Ultrawide baseline facade matching for geo-localization,''
\newblock in {\em Large-Scale Visual Geo-Localization}. 2016.

\bibitem{lin2013cross}
Tsung-Yi Lin, Serge Belongie, and James Hays,
\newblock ``Cross-view image geolocalization,''
\newblock in {\em CVPR}, 2013.

\bibitem{lin2015learning}
Tsung-Yi Lin, Yin Cui, Serge Belongie, and James Hays,
\newblock ``Learning deep representations for ground-to-aerial
  geolocalization,''
\newblock in {\em CVPR}, 2015.

\bibitem{workman2015location}
Scott Workman and Nathan Jacobs,
\newblock ``On the location dependence of convolutional neural network
  features,''
\newblock in {\em CVPR Workshops}, 2015.

\bibitem{workman2015wide}
Scott Workman, Richard Souvenir, and Nathan Jacobs,
\newblock ``Wide-area image geolocalization with aerial reference imagery,''
\newblock in {\em ICCV}, 2015.

\bibitem{vo2016localizing}
Nam~N Vo and James Hays,
\newblock ``Localizing and orienting street views using overhead imagery,''
\newblock in {\em ECCV}, 2016.

\bibitem{stumm2016robust}
Elena Stumm, Christopher Mei, Simon Lacroix, Juan Nieto, Marco Hutter, and
  Roland Siegwart,
\newblock ``Robust visual place recognition with graph kernels,''
\newblock in {\em CVPR}, 2016.

\bibitem{Zhai2017Predicting}
Menghua Zhai, Zachary Bessinger, Scott Workman, and Nathan Jacobs,
\newblock ``Predicting ground-level scene layout from aerial imagery,''
\newblock in {\em CVPR}, 2017.

\bibitem{tian2017cross}
Yicong Tian, Chen Chen, and Mubarak Shah,
\newblock ``Cross-view image matching for geo-localization in urban
  environments,''
\newblock in {\em CVPR}, 2017.

\bibitem{Hu_2018_CVPR}
Sixing Hu, Mengdan Feng, Rang M.~H. Nguyen, and Gim~Hee Lee,
\newblock ``Cvm-net: Cross-view matching network for image-based
  ground-to-aerial geo-localization,''
\newblock in {\em CVPR}, 2018.

\bibitem{sabour2017dynamic}
Sara Sabour, Nicholas Frosst, and Geoffrey~E Hinton,
\newblock ``Dynamic routing between capsules,''
\newblock in {\em NIPS}, 2017.

\bibitem{netvlad}
Relja Arandjelovic, Petr Gronat, Akihiko Torii, Tomas Pajdla, and Josef Sivic,
\newblock ``Netvlad: Cnn architecture for weakly supervised place
  recognition,''
\newblock in {\em CVPR}, 2016.

\bibitem{chopra2005learning}
Sumit Chopra, Raia Hadsell, and Yann LeCun,
\newblock ``Learning a similarity metric discriminatively, with application to
  face verification,''
\newblock in {\em CVPR}, 2005.

\bibitem{he2016deep}
Kaiming He, Xiangyu Zhang, Shaoqing Ren, and Jian Sun,
\newblock ``Deep residual learning for image recognition,''
\newblock in {\em CVPR}, 2016.

\bibitem{varior2016gated}
Rahul~Rama Varior, Mrinal Haloi, and Gang Wang,
\newblock ``Gated siamese convolutional neural network architecture for human
  re-identification,''
\newblock in {\em ECCV}, 2016.

\bibitem{schroff2015facenet}
Florian Schroff, Dmitry Kalenichenko, and James Philbin,
\newblock ``Facenet: A unified embedding for face recognition and clustering,''
\newblock in {\em CVPR}, 2015.

\bibitem{cheng2016person}
De~Cheng, Yihong Gong, Sanping Zhou, Jinjun Wang, and Nanning Zheng,
\newblock ``Person re-identification by multi-channel parts-based cnn with
  improved triplet loss function,''
\newblock in {\em CVPR}, 2016.

\bibitem{chen2017beyond}
Weihua Chen, Xiaotang Chen, Jianguo Zhang, and Kaiqi Huang,
\newblock ``Beyond triplet loss: a deep quadruplet network for person
  re-identification,''
\newblock in {\em CVPR}, 2017.

\bibitem{kinga2015method}
D~Kinga and J~Ba Adam,
\newblock ``A method for stochastic optimization,''
\newblock in {\em ICLR}, 2015.

\end{thebibliography}

\end{document}